\documentclass[11pt,a4paper]{article}
\usepackage[hyperref]{emnlp2020}
\usepackage{times}
\usepackage{linguex}
\usepackage{latexsym}
\usepackage{multirow}
\usepackage{todonotes}
\usepackage[T1]{fontenc}

\usepackage{booktabs, multirow} 
\usepackage{soul}
\usepackage{float}

\usepackage{microtype}

\usepackage[super]{nth}
\usepackage[normalem]{ulem}
\usepackage{xcolor}
\newcommand\asout{\bgroup\markoverwith
{\textcolor{purple}{\rule[.5ex]{2pt}{0.4pt}}}\ULon}

\aclfinalcopy

\title{Unsupervised Translation of German--Lower Sorbian: Exploring Training and Novel Transfer Methods on a Low-Resource Language} 

\author{Lukas Edman \qquad  Ahmet \"{U}st\"{u}n \qquad Antonio Toral \qquad Gertjan van Noord \vspace{.2cm}
 \\ Center for Language and Cognition \\ 
 University of Groningen \vspace{.1cm}
 \\ {\tt \small\{j.l.edman, a.ustun, a.toral.ruiz, g.j.m.van.noord\}@rug.nl}
}
\date{}

\begin{document}
\maketitle
\begin{abstract}
This paper describes the methods behind the systems submitted by the University of Groningen for the WMT 2021 Unsupervised Machine Translation task for German--Lower Sorbian (DE--DSB): a high-resource language to a low-resource one. Our system uses a transformer encoder-decoder architecture in which we make three changes to the standard training procedure.
First, our training focuses on two languages at a time, contrasting with a wealth of research on multilingual systems. Second, we introduce a novel method for initializing the vocabulary of an unseen language, achieving improvements of 3.2 BLEU for DE$\rightarrow$DSB and 4.0 BLEU for DSB$\rightarrow$DE. Lastly, we experiment with the order in which offline and online back-translation are used to train an unsupervised system, finding that using online back-translation first works better for DE$\rightarrow$DSB by 2.76 BLEU. Our submissions ranked first (tied with another team) for DSB$\rightarrow$DE and third for DE$\rightarrow$DSB.

\end{abstract}

\section{Introduction}

Unsupervised Neural Machine Translation (UNMT) has become increasingly useful in the field of MT, given that monolingual data is easier to gather compared to bilingual (or parallel) data. Such is especially the case for low-resource languages, which constitute the majority of languages in the world. 

The WMT 2021 Unsupervised MT Task focuses on one such low-resource language: Lower Sorbian (DSB). The task is to translate between German (DE), a high-resource language, and Lower Sorbian, which is a very low resource language with roughly 150 thousand sentences of monolingual data available for the task at hand. The unsupervised task from prior years of WMT focused on German--Czech and German--Upper Sorbian translation. Unique to this year however is the relatively small amount of monolingual data available for DSB, compared to last year in which roughly 750 thousand sentences of Upper Sorbian were available. This makes it increasingly difficult to rely on the ubiquitous state-of-the-art UNMT methods \cite{lample2019cross, song2019mass, liu2020multilingual}, as they typically rely on a large amount of monolingual data available for both languages. 

To alleviate the difficulty that comes with the lack of monolingual data for DSB, this year's WMT task allows for the use of monolingual and parallel data outside of DE--DSB. Specifically, all Upper Sorbian (HSB) data from WMT20 and all parallel data for German (DE) from WMT and OPUS \cite{tiedemann2004opus} are made available to use. Additionally, as auxiliary languages related to DSB, monolingual data for Czech (CS) and Polish (PL) is also provided.

Given the success of language transfer via multilingual models such as mBART \cite{liu2020multilingual}, this fundamentally changes this year's unsupervised task from a bilingual NMT task to a multilingual task. However, pretrained multilingual models like mBART cannot be used as they do not fit the limitations on the training data that one is allowed to use for this shared task. 

As the problem is unique to date due to the limited available data as well as the limitation on preexisting pretrained models, we aim to establish a standard for training systems under these restrictions. 
Specifically, we ask three research questions (RQs):

\begin{enumerate}
    \item Is it better to pretrain and fine-tune a multilingual model or to focus the training on a few languages at a time when data and time are limited?
    \item How can we obtain a good initialization for the vocabulary of an unseen language for which there is very limited training data?
    \item Given there are two methods for doing back-translation, online and offline, what is the best way to combine them?
\end{enumerate}

Concerning RQ1, there is a wealth of research into multilingual models, however these typically require a large amount of monolingual data for each language in addition to a wealth of computational resources and time. Therefore, we develop the hypothesis that under limited constraints (time and computational resources) training only on two languages at a time will result in better performance due to there being fewer training objectives. 

Specific to this task, we propose training on DE coupled with CS, then HSB, then DSB itself, following the order of least to most linguistically similar to the low resource language that we target: DSB. 

As for RQ2, due to the scarcity of monolingual data and complete absence of parallel data for DSB, our ability to train the model on this language is limited, thus the model's initialization for the language plays an increasingly important role in its resulting performance. Relying on the similarity of the two Sorbian languages, we aim to improve this initialization by transferring the model's knowledge of the HSB vocabulary to the DSB vocabulary. 

Finally, with respect to RQ3, \citet{garcia2020harnessing} established a method for incorporating offline and online back-translation (BT) into their MT system by first using offline BT then online BT, both following a multilingual pretraining.
However, the reverse order (i.e. online BT followed by offline BT) has not been tested to the best of our knowledge, and theoretically doing online BT should improve the quality of the synthetic data that would be used for offline BT. Therefore, we test \citeauthor{garcia2020harnessing}'s method as well as the reverse order to establish the best practice for this task specifically. 

The remaining of the paper is organized as follows.
We first outline the data we chose to use and our preprocessing steps in Section~\ref{sect:data}. We then specify our architecture and training methods in Section~\ref{sect:method}. Our results are in Section~\ref{sect:results}, followed by our conclusions in Section~\ref{sect:conclusion}.

\section{Data} \label{sect:data}

Apart from our main language pair of DE and DSB, we opted to use data from two languages that are related to the latter, namely CS and HSB.\footnote{We initially also used Polish data but did not see any improvements so we ultimately left it out.}

For HSB and DSB, we use all of the data provided for training by WMT. For DE and CS, we use data from WMT NewsCrawl, years 2010-11 and 2018-20. We chose these years as they are the most frequent years occurring in the HSB data, following~\citet{edman2020data}. For DE, we take the first 1 million sentences each from 2018-20 and 0.5 million from 2010-11, totalling 4 million. For CS, we take 0.5 million and 0.25 million from the respective years, totalling 2 million. 

In terms of parallel CS--DE data, we use MultiParaCrawl, Europarl v8, WMT News v2019, and News Commentary v16, all available from the OPUS project \cite{tiedemann2004opus}. The datasets are shown in Table \ref{tab:data}.

\begin{table}[!htp]\centering

\scriptsize
\begin{tabular}{lrrr}\toprule
Language(s) &Dataset Name & Sentences \\\midrule
CS &NewsCrawl 20{10-11, 18-20} &2,000,000 \\ \midrule
DE &NewsCrawl 20{10-11, 18-20} &4,000,000 \\ \midrule
DSB &WMT 2021 &145,198 \\ \midrule
HSB &WMT 2020 &696,271 \\ \midrule
\multirow{4}{*}{CS--DE} &Europarl v8 &568,589 \\
&MultiParaCrawl v7.1 &5,680,308 \\
&NewsCommentary v16 &204,311 \\
&WMT News v2019 &20,567 \\ \midrule
HSB--DE &WMT 2020-21 &147,521 \\
\bottomrule
\end{tabular}
\caption{Training data used in our models.}\label{tab:data}
\end{table}

While MultiParaCrawl (MPC) is the largest portion of Czech--German data, it is constructed using English as a pivot language, so we anticipate the data to be lower quality in general. As such, we run 2 models, 1 including MPC and 1 without MPC. 

For development and testing, we use the DE--HSB and DE--DSB \verb|devel| and \verb|devel_test| datasets provided by WMT. For CS--DE, we make use of the WMT News Translation Task dev set, using \verb|newstest2012| for development and \verb|newstest2013| for testing. 

All data is tokenized using the Moses toolkit \cite{Koehn:2007:MOS:1557769.1557821}. We then apply BPE \cite{sennrich-etal-2016-neural}, for all languages jointly, using FastBPE.\footnote{\url{https://github.com/glample/fastBPE}}
The segmentation is applied on the same number of randomly-selected sentences for each language, roughly 145 thousand, matching the number of sentences in the DSB training data. We experiment with the number of joins used (trying \{20, 40, 50, 60, 80\} thousand), finding 50 thousand to perform the best, according to BLEU scores after step 4.  

\section{Method} \label{sect:method}

\subsection{Architecture}
We used the MASS \cite{song2019mass} model, which is a 12-layer encoder-decoder (6 layers each) Transformer model identical to the XLM \cite{lample2019cross} architecture. The difference comes in the training, using the MASS sequence masking (MA) objective allows both the encoder and decoder to be trained in the language model pretraining phase. This can be contrasted with XLM, which only pretrains the encoder. 

\subsection{Training}
The training objectives we use are: 
\begin{itemize}
    \item MASS sequence masking (MA): Reconstructing a sentence fragment (a token sequence) given the remaining part of the sentence.
    \item Machine translation (MT): The standard translation objective with a cross-entropy loss. 
    \item Denoising auto-encoding (AE): Reconstructing the original text from a noisy version corrupted by a set of functions.  
    \item Back-translation (BT): Both online (on-the-fly) and offline back-translation, where synthetic data created from the forward direction is used to train the backward direction, and vice versa.
    \item Cross-lingual back-translation (XBT) \cite{li2020reference}: Using an intermediate reference language for forward and backward translation during online BT.
\end{itemize}

\begin{figure*}[ht]
    \centering
    \includegraphics[scale=0.45]{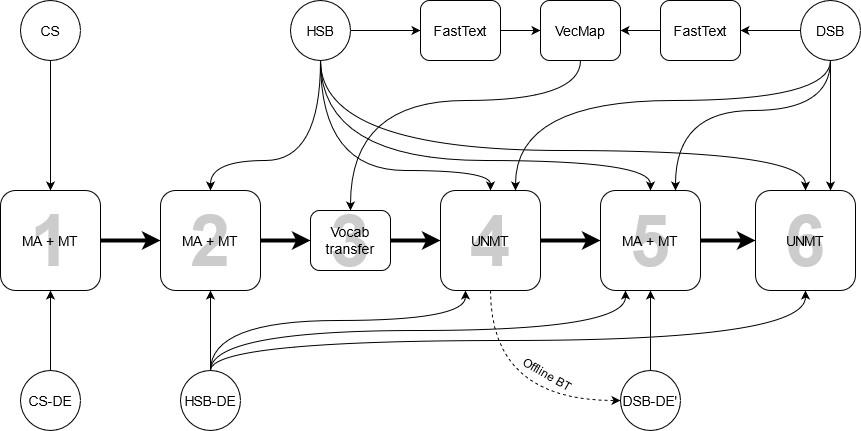}
    \caption{Diagram of the training steps. The circular nodes are datasets, the larger boxes are training steps of the MASS model, and the smaller boxes represent the steps for vocabulary transfer. German monolingual data is not shown as it is used in every step.}\label{fig:steps}
\end{figure*}

We tested various training schemes, ultimately deciding upon a 6-step process, shown in Figure \ref{fig:steps}.

In Figure \ref{fig:steps}, UNMT (steps 4 and 6) refers to the combination of AE, online BT, and XBT.
For the AE, we use word shuffling, masking, and removal as noise functions, following XLM \cite{lample2019cross}.
Online BT is done on DE--DSB as well as on HSB--DSB.\footnote{We found in initial testing that including HSB--DSB gave a slight improvement to BLEU scores.} XBT is done using our DE--HSB parallel data for the directions HSB$\rightarrow$DSB$\rightarrow$DE and DE$\rightarrow$DSB$\rightarrow$HSB.\footnote{We also tried using our DE--CS data for XBT, but this performed worse.}

\subsection{Vocabulary Transfer}
To facilitate a better alignment of DSB to DE, we make use of the linguistic similarity of DSB to HSB, coupled with the fact that HSB is expected to be reasonably well-aligned to DE after training with the HSB--DE parallel data (step 2).
As the model is language agnostic, apart from the language embeddings as well as word embeddings that occur exclusively in one language, we initially align these parts on the DSB side to the HSB side (step 3), prior to the first UNMT training (step 4).

We align the language embeddings by copying the HSB language embeddings to the DSB language embeddings. To align the vocabulary, we first train two word embeddings on the DSB and HSB data using fastText \cite{bojanowski2017enriching}. Next, we align these two embeddings using VecMap \cite{artetxe2018robust}, treating identical words in HSB and DSB as the same. 

From the aligned embeddings, we construct a bilingual dictionary. This is done by first getting the top 10 nearest HSB neighbors, according to the cosine similarities of the aligned embeddings, for each DSB word. From these 10 candidates, we choose the closest HSB word, determined by the lowest Levenshtein distance between the DSB word and the respective HSB candidate.\footnote{We strip accents before calculating the Levenshtein distance.} We also filter out DSB words that occur frequently in the DE training data, removing all those which occur more than 0.001\% of the time. 

We use this filtered bilingual dictionary to copy the embeddings within the encoder and decoder of the MASS model. Specifically, all of the embeddings for the words on the DSB side of the bilingual dictionary are copied from the embedding of their corresponding HSB word pair. 

As DSB has not yet been seen in the training, a large number of the embeddings for DSB words are essentially not learned at this stage. However, since the Sorbian languages are closely related and often the differences between words are merely in spelling, we expect this approach to help with initializing the model's DSB vocabulary. 

\begin{table}[th]\centering
\scriptsize
\begin{tabular}{lrrrr}\toprule
Data &Training Step &DE$\rightarrow$DSB &DSB$\rightarrow$DE \\\midrule
\multirow{3}{*}{Without MPC} &4 &22.46 &30.04 \\
&5 &\underline{\textbf{24.92}} &28.48 \\
&6 &23.22 &\underline{31.34} \\\midrule
\multirow{3}{*}{With MPC} &4 &23.03 &30.70 \\
&5 &\underline{23.62} &24.92 \\
&6 &22.95 &\underline{\textbf{32.06}} \\
\bottomrule
\end{tabular}
\caption{BLEU scores for our submitted models at various training steps. Best scores are in bold, and underlined scores are the models submitted to OCELoT, the submission system used for the shared task.\footnotemark \hspace{1mm}We compare performances of models trained with or without MultiParaCrawl (MPC) data. }\label{tab:bleu}
\end{table}

\subsection{Experimental Setup}
Training is done on an Nvidia V100 32GB GPU. Each training step of the model is limited to 24 hours, with the exception of our model using MultiParaCrawl data, in which the first step is trained for 2 days due to the large amount of data. The training of fastText and VecMap and application of vocabulary transfer take less than an hour, and the offline BT for DSB$\rightarrow$DE takes around 1 hour.

We use an additional stopping criterion of no improvement on the validation set in 1 million iterations. The metric we use for measuring improvement is the into-DE BLEU score, with the exception of step 6, in which we use the into-DSB BLEU score.
\protect\footnotetext{\url{https://ocelot-west-europe.azurewebsites.net/}}

The hyperparameters used follow those used in \citet{song2019mass}, except for the epoch size being set at 100 thousand steps, rather than 200 thousand steps \footnote{The implementation we use, which is based on the MASS implementation (\url{https://github.com/microsoft/MASS}), defines epochs in steps.}. We shorten this as it saves systems more often and early stopping is applied more quickly. Our code is made freely available.\footnote{{\url{https://github.com/Leukas/WMT21}}}

\begin{table}[th]\centering
\scriptsize
\begin{tabular}{lrrr}\toprule
Model &DE->DSB &DSB->DE \\\midrule
Multilingual &19.70 &25.29 \\
Multilingual + VT &20.30 &28.14 \\ 
Ours &19.25 &26.00 \\
Ours + VT &\textbf{22.46} &\textbf{30.04} \\
\bottomrule
\end{tabular}
\caption{BLEU scores comparing the two-language versus multilingual training schemes, with and without vocabulary transfer (VT).}\label{tab:mamt}
\end{table}

\section{Results} \label{sect:results}
Table \ref{tab:bleu} shows our BLEU scores for our submitted models DE--DSB, starting from step 4. As we can see, the \nth{5} step of MT using the synthetic data obtained from offline BT scores the best for DE$\rightarrow$DSB, as such we used our models at this step for our submissions for this direction. For DSB$\rightarrow$DE, we see the BLEU score actually drops for step 5, but the second phase of UNMT training improves the BLEU by roughly 1 point over step 4.

\subsection{Pretraining Two Languages at a Time}

In our first research question, we asked if it is best to train on two languages at a time. To answer this, we compare our model (without the MultiParaCrawl data) to another model trained on all 4 of our languages from the start. Specifically, we do MA for all languages, and MT for those with parallel data. We train this for 2 days, so that it is trained for the same length as steps 1 and 2. We then train using the UNMT objectives. The results are shown in Table~\ref{tab:mamt}.

Despite the model being exposed to DSB for longer, the performance is equal compared to our model without vocabulary transfer, and worse compared to the model with vocabulary transfer.

Without vocabulary transfer, the results being equal shows that exposing the model to DSB early within the training is not important to the final performance. We expect this might be due to the limited training data of DSB, as \citet{wu2020all} similarly found that the multilingual model mBERT performed more poorly on languages lacking in monolingual data. 

With vocabulary transfer, we see that the model that has not seen DSB at all benefits more from the initialization than the model which has. We believe this shows the underlying advantage of training a model on few languages at a time, as it develops a better internal representation for the auxiliary languages, which enables better transfer. Table \ref{tab:pretr} shows that the performance of our model on the auxiliary languages is better when it can focus on learning one language pair at a time.  

\begin{table}[t]\centering
\scriptsize
\begin{tabular}{lrrrrr}\toprule
Model &DE$\rightarrow$CS &CS$\rightarrow$DE &DE$\rightarrow$HSB &HSB$\rightarrow$DE \\\midrule
Multilingual &11.56 &14.36 &49.22 &49.14 \\
Ours (step 1) &\textbf{15.05} &\textbf{16.16} & & \\
Ours (step 2) & & &\textbf{52.14} &\textbf{51.57} \\
\bottomrule
\end{tabular}
\caption{BLEU scores of the auxiliary directions during pretraining.}\label{tab:pretr}
\end{table}

Moreover, while the multilingual model could learn to mimic its internal representation of HSB when encoding DSB, its representation according to Table \ref{tab:pretr} is poorer, and our model with vocabulary transfer explicitly copies the only language-dependent information the model receives, forcing an internal representation of DSB based on that of HSB. 

\subsection{Vocabulary Transfer Analysis}
We also conduct an ablation of our novel method of vocabulary transfer. Table \ref{tab:vocab_res} shows the results of step 4, with and without vocabulary transfer. We also show results for a simpler transfer method: rather than taking the top 10 most similar candidates and choosing based on Levenshtein distance, we simply select the most similar candidate.

The addition of vocabulary transfer adds over 3 BLEU to the performance. We also see both transfer methods are competitive with each other in terms of improvement to performance. We expect the simple method to work better for language pairs with less similar spelling than the Sorbian languages. However our following analysis leads us to believe the Levenshtein version may perform better for similar languages.

We also perform a form of ``zero-shot'' transfer, where we use the model from Step 2 and test its ability to translate DSB$\rightarrow$DE, despite the neural model never being trained on DSB at this stage. We contrast that with applying our 2 transfer methods. The results are in Table \ref{tab:vocab_step2}.

Without vocabulary transfer, the model expectedly has trouble with translation as it has not yet seen any DSB, so its vocabulary is not properly initialized. However with vocabulary transfer, we see an improvement of 16-18 BLEU, with the Levenshtein version performing best. This shows the degree to which a good initialization of the word embeddings can play a role in the overall performance of the model on an unseen language. Although UNMT (step 3) helps narrow the gap in performance, the difference of 3 BLEU also shows that unsupervised training can also stand to benefit from a better vocabulary initialization.   

\begin{table}[tp]\centering
\scriptsize
\begin{tabular}{lrrr}\toprule
Transfer Method &DE$\rightarrow$DSB &DSB$\rightarrow$DE \\\midrule
None &19.25 &26.00 \\
Simple &\textbf{22.64} &29.98 \\
Levenshtein &22.46 &\textbf{30.04} \\
\bottomrule
\end{tabular}
\caption{BLEU scores comparing no vocabulary transfer to our 2 methods.}\label{tab:vocab_res}
\end{table}

\begin{table}[tp]\centering
\scriptsize
\begin{tabular}{lrr}\toprule
Transfer Method &DSB$\rightarrow$DE \\\midrule
None &3.15 \\
Simple & 19.00 \\
Levenshtein &\textbf{21.27} \\
\bottomrule
\end{tabular}
\caption{Comparison of our model from step 2 with and without vocabulary transfer.}\label{tab:vocab_step2}
\end{table}

\subsection{Back-translation}
Our final research question concerns the order of back-translation. \citet{garcia2020harnessing} found that, using a multilingual model trained on a hub language (e.g. German), one can achieve noticeable improvement by first zero-shot translating into the hub language (e.g. DSB$\rightarrow$DE as done in step 5), then using this synthetic data for MT training. This can be followed by UNMT training (which includes online BT) for further improvement. We compare this method with the reverse, where we do UNMT training before offline BT, with the assumption that the better-quality translation after first training with online BT will result in better MT training. We show the results in Table \ref{tab:bt}.

As we can see, the performance of the model using only offline BT produces lower quality translations compared to using only online BT. While following offline BT with online BT makes up the difference in performance into DE, it still performs much worse into DSB. This supports our assumption that the better quality synthetic data leads to better MT training, as the main goal of creating synthetic DE data is to improve training with DSB on the target side. 

\begin{table}[t]
\centering
\scriptsize
\begin{tabular}{lrr}\toprule
Back-translation & DE$\rightarrow$DSB & DSB$\rightarrow$DE \\\midrule
Offline & 21.74 & 22.74 \\
Offline $\Rightarrow$ Online & 21.88 & \textbf{30.14} \\
Online &22.46 & 30.04 \\
Online $\Rightarrow$ Offline &\textbf{24.64} &27.89 \\
\bottomrule
\end{tabular}
\caption{BLEU scores for the different back-translation methods.}\label{tab:bt}
\end{table}

\section{Conclusion} \label{sect:conclusion}
The translation of Lower Sorbian to and from German presents a unique challenge in the field of unsupervised MT, due to the absence of parallel data and the scarcity of monolingual data for training. Therefore, the task necessitates an initial pretraining with similar, higher-resource languages. With this assumption, we experimented with various methods of pretraining, positing that training on 2 languages at a time 
is competitive with training with all languages at once, while allowing for a better initialization of DSB. 

We also showcase a new method for transferring knowledge to the word embeddings of a transformer, provided a similar language is used in pretraining. We intend to experiment with this method further to gauge its applicability for more distantly-related languages. 
Finally, the use of both online and offline back-translation can improve the performance of a model, and if not done in an iterative fashion, the order in which they are performed can greatly affect the results. 

\section*{Acknowledgments}

We would like to thank the Center for Information
Technology of the University of Groningen for their
support and for providing access to the Peregrine
high performance computing cluster.

\bibliographystyle{acl_natbib}
\bibliography{emnlp2020}

\end{document}